\documentclass[letterpaper, 10 pt, conference]{ieeeconf}  

\IEEEoverridecommandlockouts                              

\usepackage{graphics} 
\usepackage{epsfig} 
\usepackage{times} 
\usepackage{amsmath} 
\usepackage{amssymb}  
\usepackage{amsfonts}
\usepackage{bm}
\usepackage{derivative}
\usepackage[ruled]{algorithm2e}
\usepackage{url}
\usepackage{booktabs}
\usepackage[caption=false,font=normalsize,labelfont=sf,textfont=sf]{subfig}

\renewcommand{\v}[1]{\bm{#1}} 
\newcommand{\N}{\mathcal{N}} 
\newcommand{\I}{\bm{I}} 
\newcommand{\T}{\intercal} 
\newcommand{\ol}[1]{\overline{#1}} 

\newcommand{\R}{\mathbb{R}}

\title{\LARGE \bf
Incremental Learning of Probabilistic Movement Primitives (ProMPs) for Human-Robot Cooperation
}

\author{Daniel Schäle$^{1}$, Martin F. Stoelen$^{1,2}$ and Erik Kyrkjebø$^{1}$
\thanks{This work was funded by the Research Council of Norway through grant number 280771.}
\thanks{$^{1}$Department of Computer Science, Electrical Engineering and Mathematical Sciences, Faculty of Engineering and Science, Western Norway University of Applied Sciences, Førde, Norway.
        {\tt\small dasc@hvl.no}}%
\thanks{$^{2}$Centre for Robotics and Neural Systems (CRNS), University of Plymouth, Plymouth, United Kingdom.}%
}

\begin{document}

\maketitle
\thispagestyle{empty}
\pagestyle{empty}

\begin{abstract}
For a successful deployment of physical Human-Robot Cooperation (pHRC), humans need to be able to teach robots new motor skills quickly. 
Probabilistic movement primitives (ProMPs) are a promising method to encode a robot's motor skills learned from human demonstrations in pHRC settings.
However, most algorithms to learn ProMPs from human demonstrations operate in batch mode, which is not ideal in pHRC. 
In this paper we propose a new learning algorithm to learn ProMPs incrementally in pHRC settings.
Our algorithm incorporates new demonstrations sequentially as they arrive, allowing humans to observe the robot's learning progress and incrementally shape the robot's motor skill. A built in forgetting factor allows for corrective demonstrations resulting from the human's learning curve or changes in task constraints.
We compare the performance of our algorithm to existing batch ProMP algorithms on reference data generated from a pick-and-place task at our lab.
Furthermore, we show in a proof of concept study on a Franka Emika Panda how the forgetting factor allows us to adopt changes in the task.
The incremental learning algorithm presented in this paper has the potential to lead to a more intuitive learning progress and to establish a successful cooperation between human and robot faster than training in batch mode.
\end{abstract}

\section{INTRODUCTION}
Human-Robot Cooperation (HRC) has great potential for the (semi) automation of manufacturing processes with small batch sizes and frequently changing tasks. 
Due to the cognitive abilities of the human in the loop, HRC is expected to be more versatile and flexible than conventional (full) automation and could offer suitable automation concepts to small and medium sized enterprises.
For HRC to be successful in practice, the human needs to be able to teach the robot new motor skills quickly.
And in the ideal case, the human and robot are learning together at the same time as they are experimenting on how to solve the task.
\begin{figure}[thpb]
	\centering
	\includegraphics[width=\columnwidth]{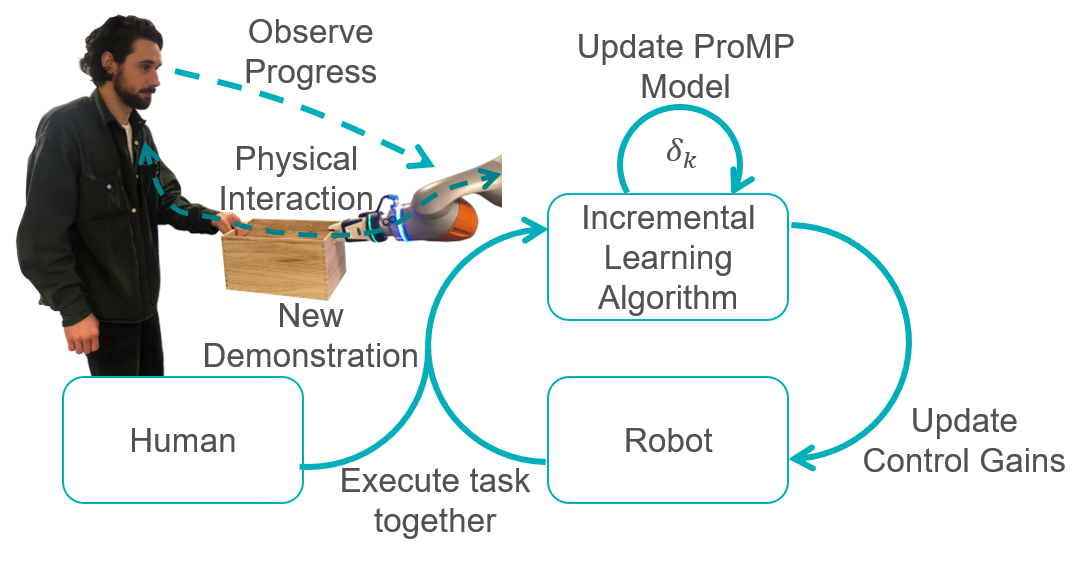}
	\caption{The intended use for our incremental learning algorithm for ProMPs in physical Human-Robot Cooperation. Human and robot learn a new task together by trying to solve the task.
	Each task execution serves as a new demonstration which is incrementally incorporated into the robot's motor skill by our algorithm.
	The human can observe the robot's learning progress and thus incrementally shape the robot's motor skill.}
	\label{fig:IncLearningScheme}
\end{figure}
A common approach within the robotics community to teach motor skills to a robot is Learning by Demonstration (LbD) in combination with movement primitives \cite{Billard.2016}.
A promising movement primitive framework are Probabilistic Movement Primitives (ProMPs) \cite{Paraschos.2013} which offer some useful features for HRC.
First, their probabilistic representation of motion allows the robot to respond appropriately to the variance inherent in human motion.
Second, the ProMP framework provides several operators to condition, combine and sequence motor skills, which makes it easier to generalize ProMPs to unknown situations which occur frequently when robots interact with humans in unstructured environments.
However, most of the algorithms in the literature to train ProMPs from human demonstrations operate in batch mode \cite{Ewerton.2015b,GomezGonzalez.2020,Paraschos.2018}, which we argue is suboptimal for HRC settings.
Batch mode means that all human demonstrations of a new task are provided at the beginning of the training. Once the desired number (batch size) of demonstrations is reached, the learning algorithm calculates the ProMP parameters, and the human and the robot can start to work on the actual task.
In batch mode, the training phase and execution phase of the task is separated, which delays the overall progress of the task.
Also, most humans will only have a vague understanding of how a robot learns new motor skills and how it reacts to certain demonstrations.
Thus, it may be difficult for lay users to provide a set of demonstrations which models the robot's desired behavior without observing any intermediate training progress.
Updating a motor skill with new demonstrations becomes increasingly difficult in batch mode since computation time and memory requirements increase with the number of demonstrations.
Hence, we claim that batch learning can not readily be used for lifelong learning with continuous shaping of motor skills.

Batch algorithms based on the expectation-maximization algorithm (EM) have been proposed by \cite{Ewerton.2015b} for maximum likelihood estimates (MLE) of single ProMPs and \cite{GomezGonzalez.2020} for maximum a posteriori estimates (MAP) of single ProMPs.
The authors in \cite{Ewerton.2015b} and \cite{GomezGonzalez.2020} treat weight vectors as hidden variables and use EM to maximize the marginal likelihood of the observed robot states with respect to the distribution parameters of the ProMP. 
The main difference between \cite{Ewerton.2015b} and \cite{GomezGonzalez.2020} is that \cite{GomezGonzalez.2020} regularize their parameter estimates by means of a prior distribution over the ProMP parameters.
The general advantage of the approach in \cite{Ewerton.2015b} and \cite{GomezGonzalez.2020} over algorithms that compute the weights by least squares (such as \cite{Paraschos.2018, Koert.2018, Conkey.2019}) is the increased robustness against noisy and incomplete demonstration data \cite{Ewerton.2015b, GomezGonzalez.2020}.
Apart from that, offers the prior parameter distribution in \cite{GomezGonzalez.2020} a convenient way of regularization and to incorporate prior assumptions about the ProMP parameters. Both is important for incremental learning, due to the sparse data at the beginning of training.

An incremental learning algorithm for a Gaussian mixture model of Interaction ProMPs (IProMPs) was proposed by \cite{Koert.2018}.
IProMPs are a variant of ProMPs which are learned from demonstrations containing a robot trajectory and a corresponding human trajectory recorded via a motion capturing system. The resulting joint model, the IProMP, is used to make the robot respond appropriately to new human motion inputs in an interaction scenario.
In case of a new demonstration, \cite{Koert.2018} compute a point estimate of the weight vector via least squares and then update the distribution parameters of the Gaussian mixture model with an EM algorithm based on \cite{Engel.2010}.
In \cite{Lee.2011}, an incremental learning algorithm with a forgetting factor is presented. However, the movement primitives in \cite{Lee.2011} are different from ProMPs and based on Hidden Markov Models.

In this paper, we want to overcome the aforementioned limitations of batch learning with an incremental learning algorithm for ProMPs which allows human and robot to jointly learn new motor skills in pHRC settings. 
The intended use of our incremental learning algorithm in physical Human-Robot Cooperation is shown in Fig.~\ref{fig:IncLearningScheme}.
The driving factors behind such an algorithm are to enable a successful cooperation from the first demonstration onwards, and to create an intuitive and open-ended learning progress.
Our algorithm incorporates new demonstrations sequentially as they arrive, allowing the human to incrementally shape the robot's motor skill.
Thus, the user can observe how the robot reacts and adapts its behavior to each demonstration and how well the robot performs already with respect to the task objective.
In a task new to both human and robot, the human will be subject to a learning curve as well. We allow for improvements of the human's demonstrations by introducing a forgetting factor into our algorithm \cite{Sato.2000}, which discounts the influence of older demonstrations to the ProMP. Furthermore, the forgetting factor makes it possible to adapt a motor skill to changes in the task constraints in later stages of the cooperation.

For our purpose, none of the training algorithms found in the literature is optimal. We want to profit from the robustness of the two batch algorithms, while on the other hand, the pHRC settings under our consideration require an incremental and open-ended training progress. 
Furthermore, we do not want to use IProMPs, since we want to avoid using a motion capturing system and explore a cooperative learning process based only on the physical interaction between human and robot.
In this paper, we focus on learning single ProMPs instead of libraries in terms of mixture models.

\section{INCREMENTAL LEARNING}
In this section we introduce our algorithm for the incremental learning of ProMPs in pHRC settings: First, we review relevant aspects of the ProMP framework \cite{Paraschos.2013}, and second, we describe how we use an online variant of an expectation maximization algorithm to learn ProMP parameters from human demonstration.

\subsection{Probabilistic Movement Primitives} 
A ProMP represents a distribution over trajectories \cite{Paraschos.2013}. 
A trajectory $ \v{\tau} = \{\v{y}_t\}_{t=1}^T $ is a time-series of vector-valued robot states $ \v{y}_t \in \mathcal{S} $ in a state space $ \mathcal{S} \subseteq \R^D $, where $ D $ is the dimension of the state space. Both joint space and task space are valid choices for $ \mathcal{S} $, and the choice depends on the desired application.\\
To reduce the number of model parameters, trajectories are concisely represented as weight vectors in a basis function model.
A weight vector $ \v{w} \in \R^{KD} $ is related at time $ t $ to the robot's state $ \v{y}_t $ through a time dependent, block diagonal basis function matrix $ \v{\Phi}_t \in \R^{D \times KD} $.
The matrix 
\begin{equation}
	\v{\Phi}_t = 
	\begin{bmatrix}
		\v{\phi}_{1,t}^\T & \cdots & \v{0} \\
		\vdots & \ddots & \vdots \\
		\v{0} & \cdots & \v{\phi}_{D,t}^\T
	\end{bmatrix}	
\end{equation}
contains on its diagonal a row vector $ \v{\phi}_{d,t}^\T \in \R^{K} $ for each degree of freedom, which again contains the values of $ K $ normalized, evenly spaced, Gaussian basis functions $ \phi_k(t) $ evaluated at time $ t $.
The basis function model has the form
\begin{equation}
	\v{y}_t = \v{\Phi}_t \v{w} + \v{\epsilon}_y \ .
\end{equation}
The weight vector $ \v{w} $ is a vertical concatenation of $ D $ column vectors $ \v{w}_d \in \R^K $, representing the weight vectors of each individual degree of freedom of the robot.
The last term $ \v{\epsilon}_y \in \R^D $ is a vector containing the observation noise which is assumed to be independent and identically distributed and to follow the normal distribution $ \N(\v{0}, \v{\Sigma}_y) $.\\
Given a weight vector $ \v{w} $, it follows that a trajectory $ \v{\tau} $ consisting of $ T $ time steps is distributed according to
\begin{equation} \label{eq:tau_cond_w}
	p(\v{\tau}|\v{w}) = \prod_{t=1}^{T} \N(\v{y}_t|\v{\Phi}_t \v{w}, \v{\Sigma}_y) \ .
\end{equation}
Multiple demonstrations of the same movement are expected to differ slightly.
This implies that different weight vectors $ \v{w}_n $ are needed to represent the $ n $ different instances of a movement.
The underlying mechanism generating the weight vector samples is assumed to be a Gaussian distribution
\begin{equation} \label{eq:w_cond_theta}
	p(\v{w}|\v{\theta}_w) = \N(\v{w}|\v{\mu}_w, \v{\Sigma}_w) \ ,
\end{equation}
where $ \v{\theta}_w = \{ \v{\mu}_w, \v{\Sigma}_w \} $ are the distribution parameters.
The mean vector $ \v{\mu}_w \in \R^{KD} $ summarizes the mean of the demonstrations in each degree of freedom.
The covariance matrix $ \v{\Sigma}_w \in \R^{KD \times KD} $ represents the variances and covariances of the demonstrations in respectively between each degree of freedom.
Learning a ProMP from demonstration requires to find the distribution parameters $ \v{\theta}_w $ that explain the demonstration data best.
It is also possible to replace the normal distribution $ \N(\v{w}|\v{\mu}_w, \v{\Sigma}_w) $ with other types of probability distributions such as mixture models. Gaussian mixture models (GMMs) have been used on several occasions to model mixtures of motor skills with ProMPs \cite{Koert.2018, Conkey.2019, Maeda.2017, Ewerton.2015}.
	
The complete ProMP model is obtained by combining Eq.~\ref{eq:tau_cond_w} and~\ref{eq:w_cond_theta} and subsequently marginalizing out the weights $ \v{w} $. A movement encoded as a ProMP with parameters $ \v{\theta}_w = \{ \v{\mu}_w, \v{\Sigma}_w \} $, is represented by the distribution
\begin{equation}\label{eq:promp_model}
	p(\v{\tau}|\v{\theta}_w) = \int \N(\v{w}|\v{\mu}_w, \v{\Sigma}_w) \prod_{t=1}^{T} \N(\v{y}_t|\v{\Phi}_t \v{w}, \v{\Sigma}_y) d\v{w} \ .
\end{equation}
The time signals of the demonstrations are normalized to compensate for different durations.
For this purpose, the time signal $ t $ is replaced by a phase variable $ z_t $ that ensures that every demonstration runs between $ z_0 = 0 $ and $ z_T = 1 $.

\subsection{Incremental Learning Algorithm}
In this paper, we propose to use an online variant of the EM algorithm known as stepwise EM (sEM) \cite{Sato.2000, Liang.2009} to train ProMPs incrementally in pHRC settings.
As usual in EM, the objective of the algorithm is to maximize the likelihood of the observed data $ \v{Y} = \{\v{\tau}_n\}_{n=1}^N $, consisting of $ N $ trajectories $ \v{\tau}_n $, with respect to the model parameters $ \v{\theta}_w $. The weight vectors $ \v{w}_n $ are treated as hidden variables.
The marginal likelihood is given by
\begin{equation}\label{eq:obs_data_likelihood}
	p(\v{Y}|\v{\theta}_w) = \prod_{n=1}^{N} \int p(\v{w}_n|\v{\mu}_w,\v{\Sigma}_w)  	\prod_{t=1}^{T} p(\v{y}_{nt}|\v{w}_n) d\v{w}_n \ .
\end{equation}

For an effective use of the sEM algorithm we exploit the properties of exponential family distributions.
The joint distribution of observed and hidden variables $ p(\v{\tau}, \v{w}|\v{\theta}_w) $ for a single trajectory $ \v{\tau} $ is an exponential family with sufficient statistics
$ \v{s}_1 = \v{w} $, 
$ \v{s}_2 = \v{w} \v{w}^\T $ and 
$ \v{s}_3 = \sum_{t=1}^{T} \v{y}_{t} \v{y}_{t}^\T -2  \v{y}_{t} \v{w}^\T \v{\Phi}_{t}^\T + \v{\Phi}_{t} \v{w} \v{w}^\T \v{\Phi}_{t}^\T $.
The sufficient statistics $ \v{s} $ offer a convenient way to summarize arbitrary amounts of demonstrations without loss of information.
This is a useful property for online learning in terms of EM, since it means that we can compute the expected sufficient statistics (ESS) in the E-step, accumulate them in some form as new demonstrations arrive, and then compute the MLE or MAP of the ProMP parameters from the accumulated ESS in the M-step.

\begin{algorithm}[thpb]
	\DontPrintSemicolon
	\LinesNumbered 
	\caption{Stepwise EM Algorithm for Training ProMPs incrementally in pHRC settings. \label{alg:Step_EM}} 
	\KwData{A new demonstration $ \v{\tau} = \{\v{y}_t, z_{t}\}_{t=1}^{T'} $ containing the robot states $ \v{y}_t $ and corresponding normalized time stamps $ z_{t} $.}
	\KwIn{Step size reduction power $ 0.5 < \beta \leq 1 $, initial values for $ \v{\mu}_w, \v{\Sigma}_w,\v{\Sigma}_y $. Typically $ \v{\mu}_w = \v{0}$, $\v{\Sigma}_w = \v{I}$, $\v{\Sigma}_y = \v{I} $.}
	\KwOut{ProMP parameters $ \v{\mu}_w, \v{\Sigma}_w, \v{\Sigma}_y $.} 
	\BlankLine
	
	Initialize $ \v{u} = \v{0} $, $ \eta = 0 $, $ T = 0 $, $ N = 1 $\;
	Compute initial step size $ \delta_N \gets (N + 1)^{-\beta} $\;
	
	\If{new data $ \v{\tau} $ available}{	
		$ \v{\Phi}_t \gets \v{\Phi}(z_{t}) \ \forall t $\;
		$ \v{S}_w \gets \Bigl(\v{\Sigma}_w^{-1} + \sum_{t=1}^{T} \v{\Phi}_{t}^\T \v{\Sigma}_y^{-1} \v{\Phi}_{t} \Bigr)^{-1}$ \label{algline:Step_EM_S_w}\;
		$ \overline{\v{w}} \gets \v{S}_w \Bigl(\v{\Sigma}_w^{-1} \v{\mu}_w + \sum_{t=1}^{T} \v{\Phi}_{t}^\T \v{\Sigma}_y^{-1} \v{y}_{t} \Bigr) $ \label{algline:Step_EM_w_new}\;
		
		$ \v{u}'_1 \gets \ol{\v{w}} $ \label{algline:Step_EM_ESS_begin}\;
		$ \v{u}'_2 \gets \ol{\v{w}}\ol{\v{w}}^\T + \v{S}_w $\;
		$ \v{u}'_3 \gets \sum_{t=1}^{T'} \v{y}_{t} \v{y}_{t}^\T - 2 \v{y}_{t} \ol{\v{w}}^\T \v{\Phi}_{t}^\T + \v{\Phi}_{t} (\ol{\v{w}}\ol{\v{w}}^\T + \v{S}_w ) \v{\Phi}_{t}^\T $	\label{algline:Step_EM_ESS_end}\;
		
		$ \v{u} \gets (1 - \delta_N) \v{u} + \delta_N \v{u}' $  \label{algline:Step_EM_ESS_interpolation}\;
		
		$ \eta \gets (1 - \delta_N) \eta + \delta_N $\;
		$ T \gets (1 - \delta_N) T + \delta_N T' $\;
		
		$ \v{\mu}_w^* \gets \frac{1}{\eta} \v{u}_1 $ \label{algline:M_step_start} \;
		$ \v{\mu}_w \gets \frac{1}{N + k_0} (k_0 \v{m_0} + N \v{\mu}_w^*) $ \label{algline:M_step_map_mu} \;
		$ \v{\Sigma}_w^* \gets \frac{1}{\eta}  \v{u}_2 - \v{\mu}_w \v{\mu}_w^\T $\; 

		$ \v{\Sigma}_w \gets \frac{\v{S}_0 + N \v{\Sigma}_w^* + \tfrac{k_0 N}{k_0 + N} (\v{\mu}_w^* - \v{m}_0)(\v{\mu}_w^* - \v{m}_0)^\T}{N + v_0 + KD +2}  $\label{algline:M_step_map_sigma} \;
		$ \v{\Sigma}_y \gets \frac{1}{T} \v{u}_3 $ \label{algline:M_step_end}\;
		
		$ N \gets N + 1 $\;
		$ \delta_N \gets (N + 1)^{-\beta} $\;
		\Return $ \v{\mu}_w, \v{\Sigma}_w, \v{\Sigma}_y $
	}	
\end{algorithm} 

The pseudo code for the proposed algorithm is shown in Algorithm \ref{alg:Step_EM}.
The ESS $ \v{u}' $ are computed as the expected values of the sufficient statistics $ \v{s} $ based on the posterior distribution over the hidden variables $ p(\v{w}|\v{\tau},\v{\theta}_w^{old}) $ given the current model parameters $ \v{\theta}_w^{old} $.
The posterior distribution over the hidden variables is computed in line \ref{algline:Step_EM_S_w} and \ref{algline:Step_EM_w_new}; the ESS in line \ref{algline:Step_EM_ESS_begin} to \ref{algline:Step_EM_ESS_end}.
In sEM, whenever a new demonstration is added, the ESS are computed solely for the new demonstration. Previously added demonstrations are not stored and visited again. For comparison: in each iteration of batch EM the computations in the E-step are performed for the entire data set.
Since the ESS for a single demonstration would be a poor approximation to the ESS across the complete data set, the sEM algorithm interpolates between the sum of all previous statistics $ \v{u} $ and the statistics of the latest demonstration $ \v{u}' $. This interpolation is done in line \ref{algline:Step_EM_ESS_interpolation} in Algorithm \ref{alg:Step_EM}.
The resulting weighted sum of old and new statistics is then used to update the model parameters $ \v{\theta}_w $ in the M-step.
For the MAP estimation in the M-Step in line \ref{algline:M_step_map_mu} and \ref{algline:M_step_map_sigma} we use a normal-inverse-Wishart prior, the conjugate prior of the multivariate Gaussian distribution $ p(\v{w}|\v{\theta}_w) $. See \cite{GomezGonzalez.2020} for a detailed description.

The interpolation in line \ref{algline:Step_EM_ESS_interpolation} is governed by a step size $ \delta_N $, which controls the strength of the sum of the previous statistics $ \v{u} $ over the new statistics $ \v{u}' $. 
Similar to \cite{Liang.2009}, we use a step size $ \delta_N = (N + 1)^{-\beta} \text{ where } 0.5 < \beta \leq 1 $, which decays with the number of demonstrations or parameter updates $ N $. The user-defined parameter $ \beta $ is used to tune the decay of the step size.
A smaller $ \beta $ will lead to larger, slowly decaying step sizes $ \delta_N $ and hence, to larger updates to the statistics $ \v{u} $.

Note that all computations can be performed without storing the previous demonstrations and their ESS, which means that the memory use of the algorithm is constant.\\
The algorithm can be modified to train on mini-batches instead of single demonstrations \cite{Liang.2009}. Training in mini-batches has the advantage to produce more stable updates but slows down the incremental training progress by pooling a few demonstrations before updating the ProMP parameters. We have omitted the mini-batches here for brevity.

The step size $ \delta_N $ serves as a forgetting factor in our learning algorithm. More precisely, the step size weighs the contribution of ESS computed several iterations ago less. This is important in two regards for the incremental learning setting under our consideration: 
First, from a probabilistic perspective, it has to be kept in mind that in each iteration the ESS are computed using the posterior distribution over the hidden variables under the \emph{current} ProMP parameters. It follows that the posterior distributions, and hence the ESS, computed in the first few iterations are potentially far away from their true value, since they were computed based on crude estimates of the model parameters. 
It is crucial to forget those crude early estimates during the training, such that the ESS computed after a number of updates to the model parameters quickly dominate the ESS from early iterations and the estimated model parameters approach their true values faster. 
Second, from a motor learning perspective, discounting the contributions from early demonstrations to a motor skill is important. As described earlier, the quality of the human demonstration is expected to increase as the learning of a new motor skill proceeds. Apart from that, forgetting demonstrations which where provided a long time ago helps to adapt a motor skill to gradual changes in the task structure. Note that for an open-ended shaping of a motor skill, the step size should approach a value greater than zero.\\

\section{EXPERIMENTAL EVALUATION}\label{sec:Experiments}
\begin{figure}[thbp]
	\centering
	\includegraphics[width=1\linewidth]{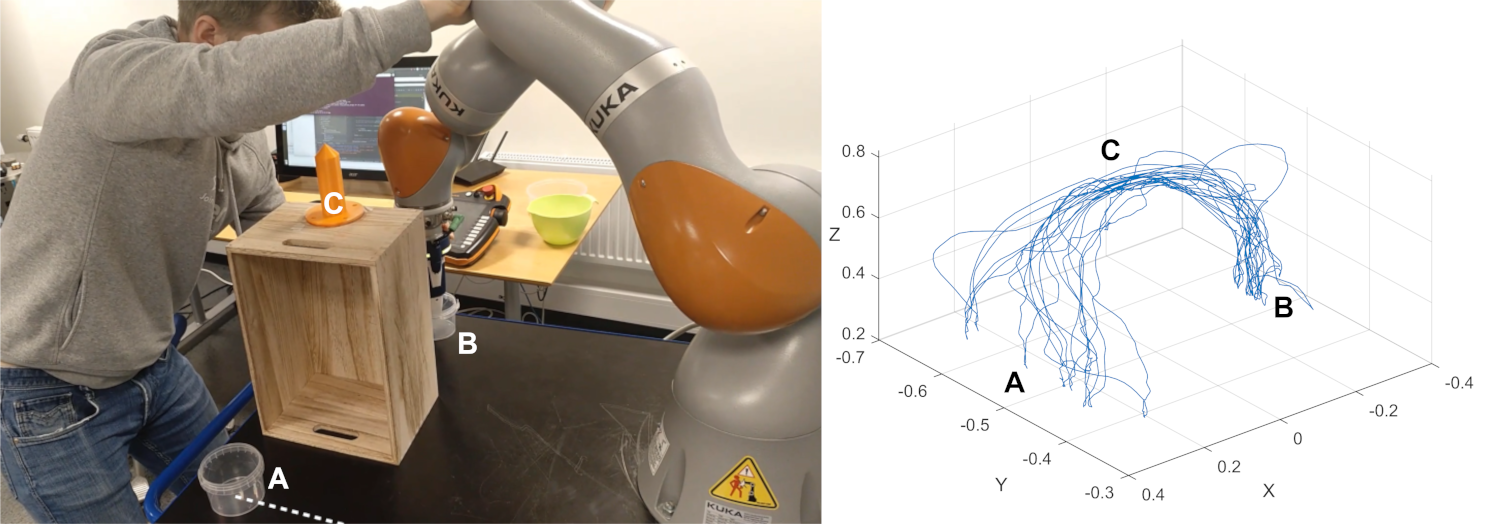}
	\caption{Pick and place task used to generate a reference ProMP. The task objective was to move the robot's end effector between cup A and B, while passing the orange tip C precisely with the gripper fingers. To introduce variance in the task constraints, Cup A was moved along the dotted line during the trials. \textbf{Left:} Task setting with KUKA iiwa 14 manipulator. \textbf{Right:} Recorded end-effector trajectories in task space.}
	\label{fig:lbdcomb4}
\end{figure}

\begin{figure}[tbp]
	\centering			
	\subfloat{\includegraphics[width=0.5\linewidth]{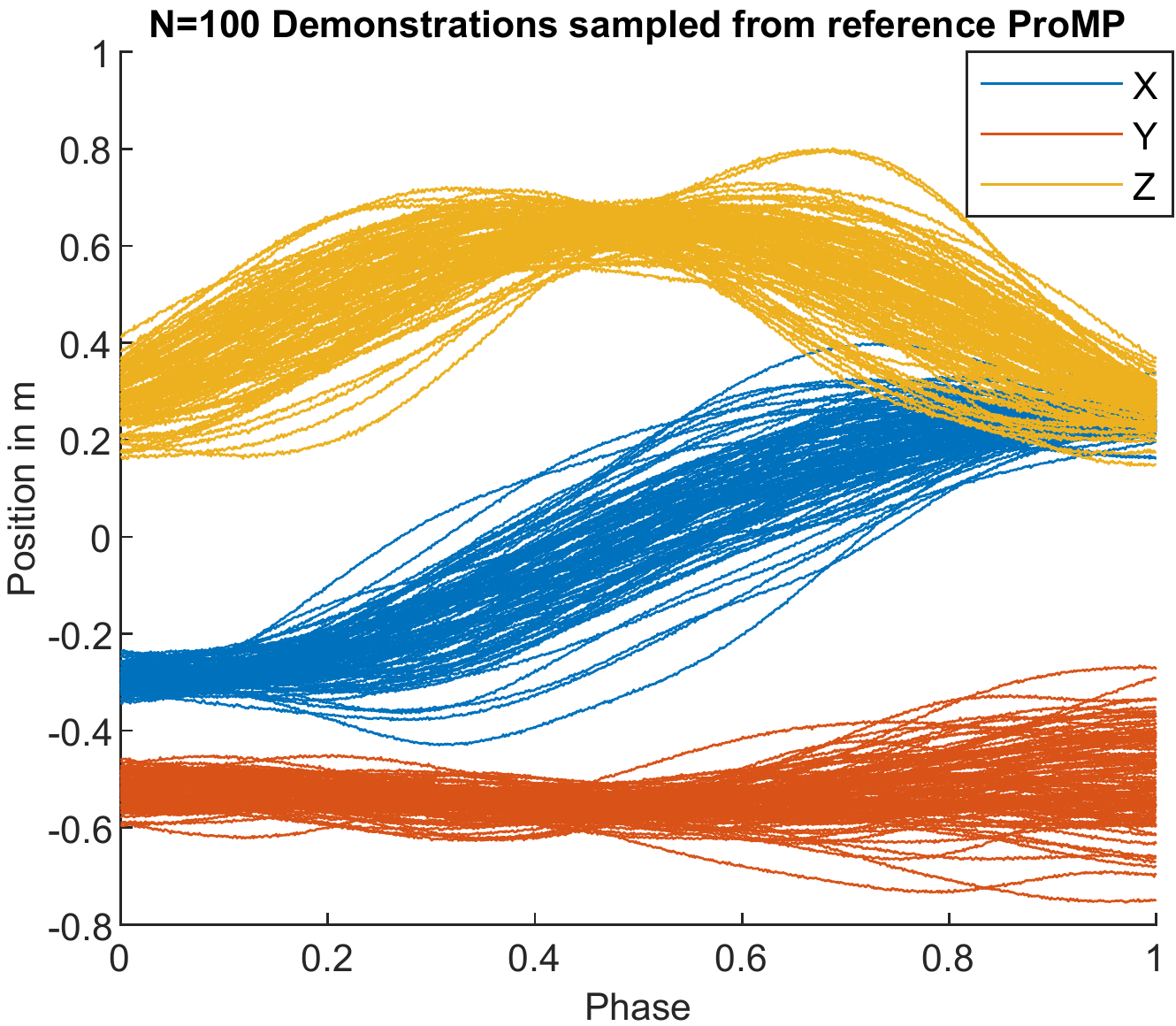}}
	\hfill
	\subfloat{\includegraphics[width=0.5\linewidth]{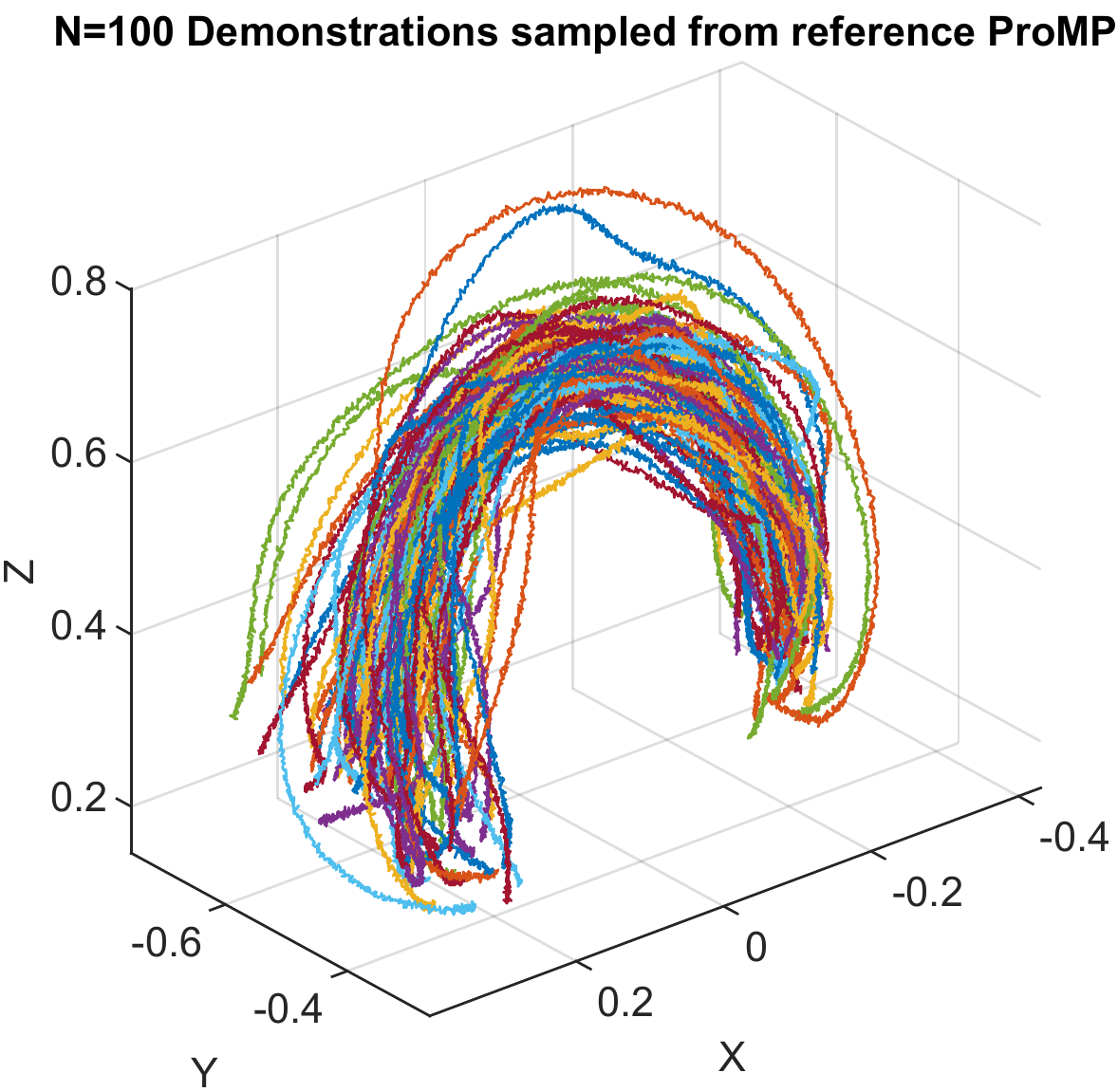}}
	\caption{The training data set used in our experiments: $ N=100 $ Demonstrations sampled from a reference ProMP. \textbf{Left:} X,Y and Z components of the sampled demonstrations plotted over the movement phase $ z_t $. \textbf{Right:} sampled demonstrations shown in task space.}
	\label{fig:sampleddemos}
\end{figure}

For the experimental evaluation of our algorithm we generated a reference ProMP based on data of a pick-and-place task done with a KUKA LBR iiwa 14 R820 robot. The task setting and recorded end-effector trajectories in task space are shown in Fig.~\ref{fig:lbdcomb4}.
The reference ProMP has $ D = 3 $ DOF and uses $ K = 10 $ basis functions, yielding a mean vector $ \v{\mu}_w^{ref} \in \R^{30} $, a covariance matrix $ \v{\Sigma}_w^{ref} \in \R^{30\times30} $ and a observation noise covariance matrix $ \v{\Sigma}_y^{ref} \in \R^{3\times3} $. We generated the reference parameters based on the empirical distribution of the robot data downsampled to $ K $ time steps. The observation noise was sampled from a Wishart distribution. 
We sampled $ N=100 $ demonstrations from this reference ProMP to generate the training data set shown in Fig.~\ref{fig:sampleddemos}, and used in the following experiments.

\subsection{Comparison of Training Algorithms}
\label{sec:ComparisonAlgorithms}
In the first experiment, we compare the performance of different training algorithms for ProMPs from literature with the incremental algorithm presented in this paper.
The performance of the different algorithms can be assessed by comparing the estimated parameters to the reference.
For a quantitative quality measure of the estimated parameters we use the Bhattacharyya distance $ D_B $ to compute the distance between the distribution of the reference ProMP to the ProMP distribution of the algorithm tested.
Also, we look at the matrix condition number $ \kappa $ of the covariance matrix $ \v{\Sigma}_w $ to asses the numerical stability of the estimated covariance matrices. The parameter $ \kappa $ is defined as the ratio of the largest singular value of $ \v{\Sigma}_w $ to the smallest. A large condition number indicates a nearly singular matrix, which can not be inverted reliably. Estimating full covariance matrices is generally difficult, especially with a low number of data/demonstrations in relation to the matrix size. However, invertible covariance matrices are crucial for a number of operations in the ProMP framework; thus, numerical stability of the parameter estimates are a relevant performance measure for the training algorithms. 
The different training algorithms
MLE with Ridge Regression \cite{Paraschos.2018},
MLE with (batch) EM \cite{Ewerton.2015b},
MAP with (batch) EM \cite{GomezGonzalez.2020} and
the MAP with sEM proposed in this work
were applied to the training data set. 
The specific parameter settings are given in Table~\ref{tab:DistancesToRef}.
For the MAP estimation in our sEM algorithm we used the same parameters for the normal-inverse-Wishart prior as the authors in \cite{GomezGonzalez.2020}: $ \v{m}_0 = \v{0} $, $ k_0 = 0 $, $ v_0 = KD + 1 $ and $ \v{S}_0 = (v_0 + KD + 1) \:\text{blockdiag}(\v{\Sigma}_w^*) $.\\  
In the last row of Table~\ref{tab:DistancesToRef}, sEM is used as a batch algorithm, by letting it pass five times incrementally over the data. This additional test was done to compare our sEM algorithm to the EM algorithms on condition of visiting each demonstration equally often. Note that sEM in the online setting (one pass over the data) only visits the $ 100 $ demonstrations once, while the EM-based batch algorithms perform five iterations, hence visit each demonstration five times. Using our incremental algorithm in batch mode is of course contrary to the rationale of this paper, but legitimate for a fair theoretical comparison to the batch EM algorithms.

The resulting performance measures $ D_B $ and $ \log\kappa(\v{\Sigma}_w) $ after applying the training algorithms are shown in Table~\ref{tab:DistancesToRef}. 
MAP estimation with EM performs best with respect to both Bhattacharyya distance $ D_B $ and matrix condition number $ \kappa $. Our algorithm, MAP with sEM, estimates in batch mode (5 passes) as well as online mode covariance matrices with condition numbers comparable to those of MAP with EM. In online mode, our algorithm has the greatest of all Bhattacharyya distances, though the distance is still of comparable magnitude to the other ones. However, with five passes over the data, our algorithm yields the second smallest distance and comes quite close to MAP with sEM.
\begin{table}
	\caption{Comparison of the training algorithms.}
	\begin{tabular}{@{}lllr@{}}
		\toprule
		& 														& \multicolumn{2}{c}{Performance Measure}	\\
		\cmidrule{3-4} 
		Algorithm 				& Algorithm settings			& $ D_B $  & $ \log\kappa(\v{\Sigma}_w) $   \\
		\midrule
		Reference 				& -								&  - 	   & 4.1988 \\
		MLE with Ridge Reg.		& $ \lambda = 10^{-12} $		&  0.8077  & 5.6010 \\
		MLE with EM 			& 5 Iterations					&  0.8098  & 5.6046 \\
		MAP with EM 			& 5 Iterations					&  0.6787  & 4.8632 \\
		MAP with sEM 			& $ \beta = 0.75 $				&  0.8978  & 4.9552 \\
		MAP with sEM 			& $ \beta = 0.75 $, 5 Passes	&  0.7012  & 4.9325 \\
		\bottomrule
	\end{tabular}
	\label{tab:DistancesToRef}
\end{table}

\subsection{Incremental Training Progress}\label{ssec:ExpIncTrainProg}
In our second experiment, we inspect the incremental training progress of our online algorithm. It is critical to ensure that the incremental parameter estimates are reasonable and usable.
We apply our algorithm in online-mode (1 pass over the data), with a step size reduction power $ \beta = 0.75 $ on the same training data set as before. Hence, we can once more compare the estimated ProMP parameters to those of the reference ProMP. Again, we use the same prior parameters as in \cite{GomezGonzalez.2020} for MAP estimation.

We use different indicators to analyze the parameter estimates during the training. 
As before, we use the Bhattacharyya distance $ D_B $ to measure the distance to the distribution of the reference ProMP.
To analyze the mean $ \v{\mu}_w $ and covariance $ \v{\Sigma}_w $ independently, we compute the relative errors of the Frobenius norms of the two parameters. 
The Frobenius norm $ \|\v{A}\|_F $ of a matrix $ \v{A} \in \R^{m \times n} $ and the relative error $ E_F $ are defined as
\begin{align}\label{eq:FrobeniusNorm}
	\|\v{A}\|_F := \sqrt{\sum_{i=1}^{m}\sum_{j=1}^{n}|a_{ij}|^2} \ , && E_F = \frac{\|\v{A_{ref}}-\v{A}\|_F}{\|\v{A_{ref}}\|_F} \ .
\end{align}
To assess the numerical stability of the covariance estimates, we compute the logarithm of the matrix condition number of $ \v{\Sigma}_w $.
Further, we compute the rotation angle between the first principal component of the ProMP distribution before and after adding a new demonstration. Large rotations of the first principal component indicate major changes in the covariance structure, either caused by an ill-conditioned covariance matrix that is sensitive to variations in the input data, or by large actual changes in the demonstration data.

\begin{figure}[tbp]
	\centering
	\subfloat{\includegraphics[width=0.5\linewidth]{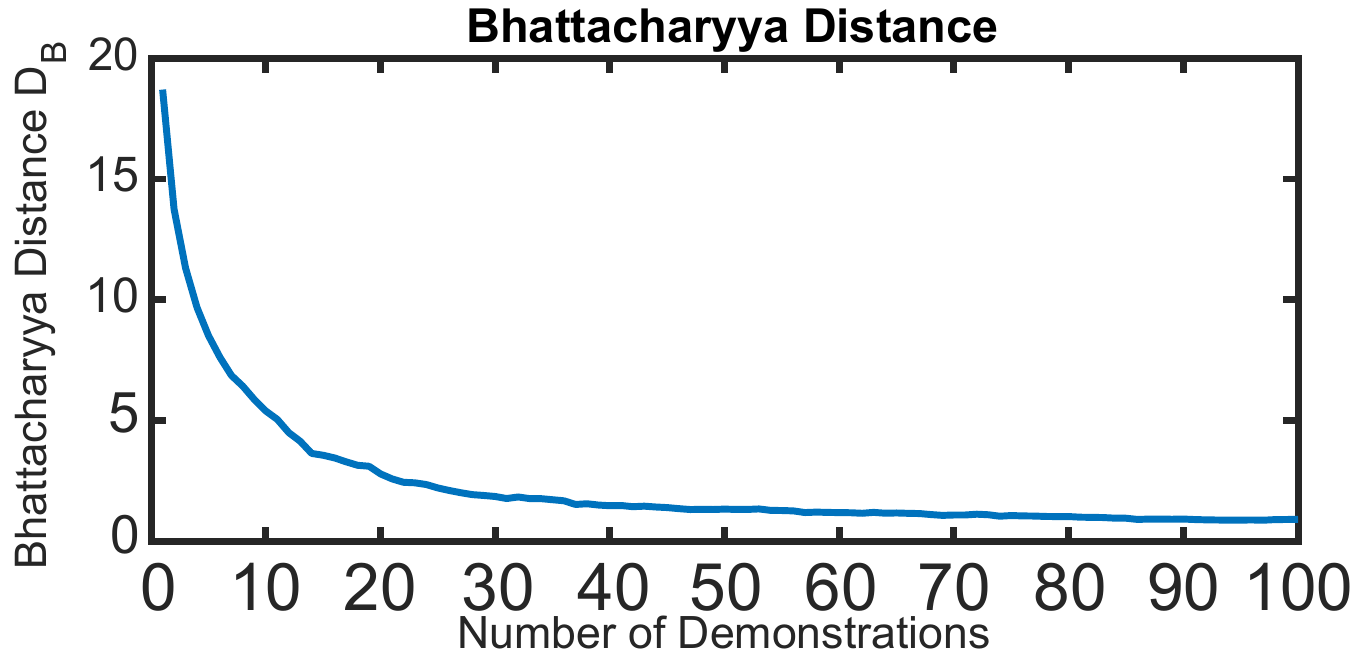}}
	\subfloat{\includegraphics[width=0.5\linewidth]{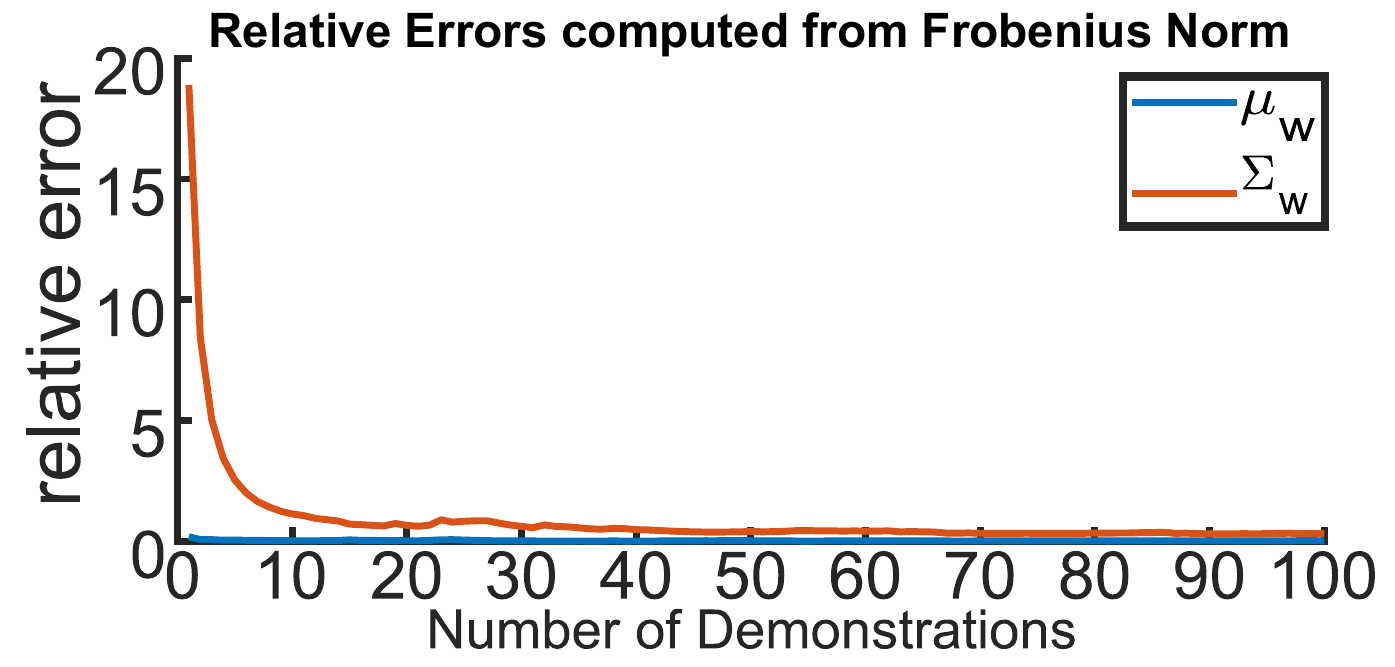}}
	
	\subfloat{\includegraphics[width=0.5\linewidth]{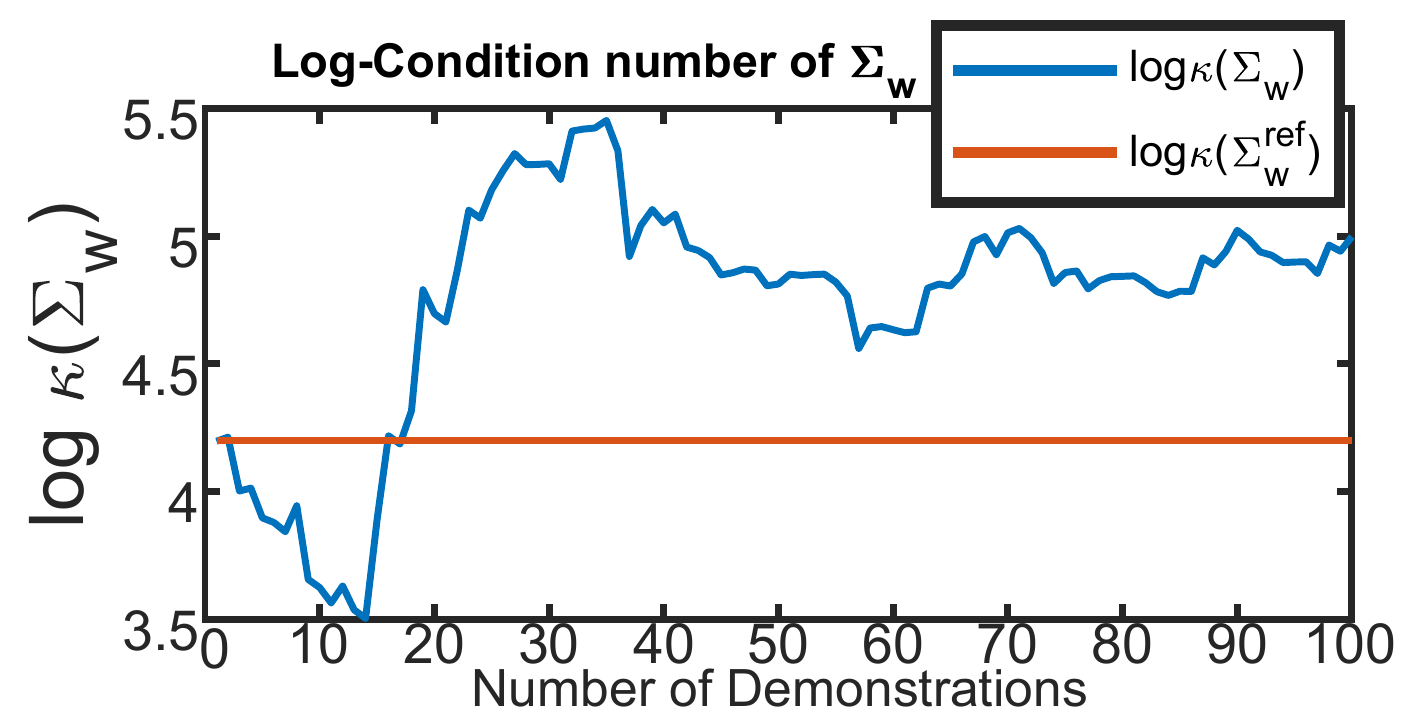}}
	\subfloat{\includegraphics[width=0.5\linewidth]{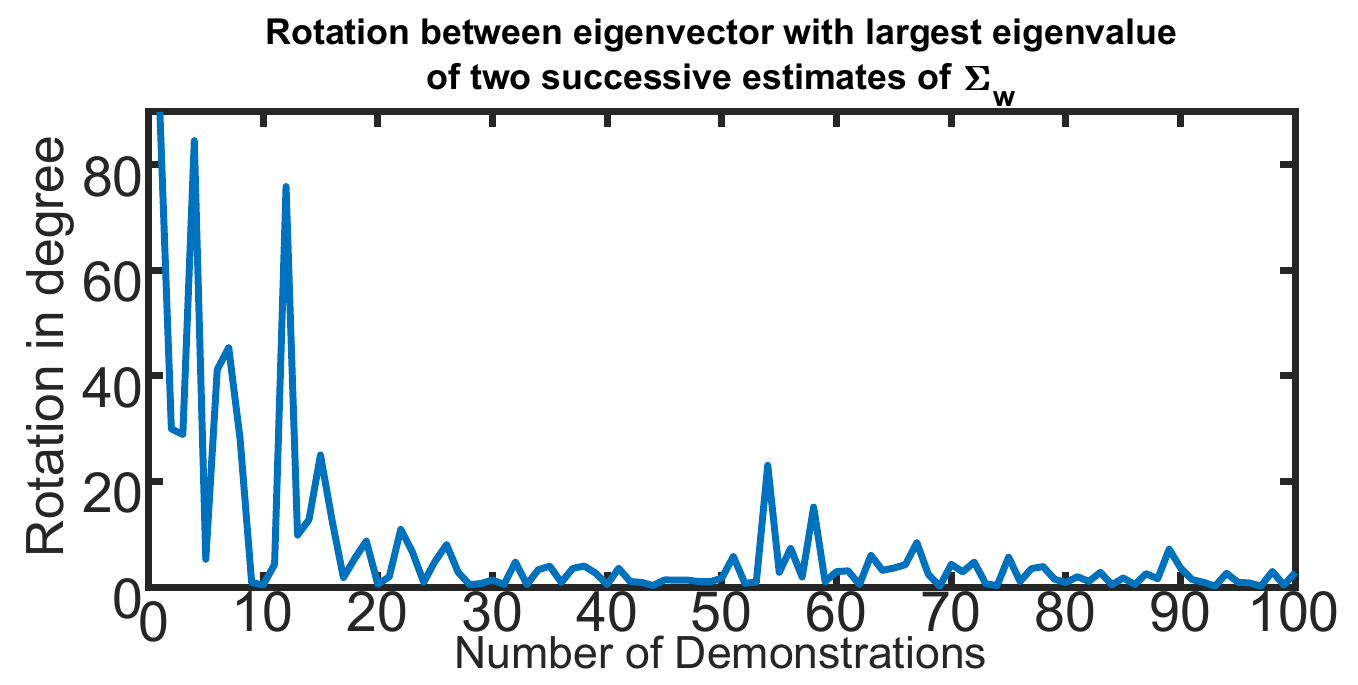}}
	\caption[Incremental training progress.]{Incremental training progress with sEM. \textbf{Top Left:} Bhattacharyya distance $ D_B $ to distribution of reference ProMP. \textbf{Top Right:} Relative Errors of the Frobenius norm $ E_F $ of the estimated ProMP parameters to the reference. \textbf{Bottom Left:} Logarithm of the matrix condition number of the estimated covariance matrix $ \v{\Sigma}_w $. \textbf{Bottom Right:} Rotation of the eigenvector with the highest corresponding eigenvalue (first principal component) between to two successive estimates of $ \v{\Sigma}_w $. Large rotations indicate major changes in the variance structure of the ProMP distribution.}
	\label{fig:incProgMetrics}
\end{figure}

\begin{figure*}[!t]
	\centering
	\includegraphics[width=\textwidth]{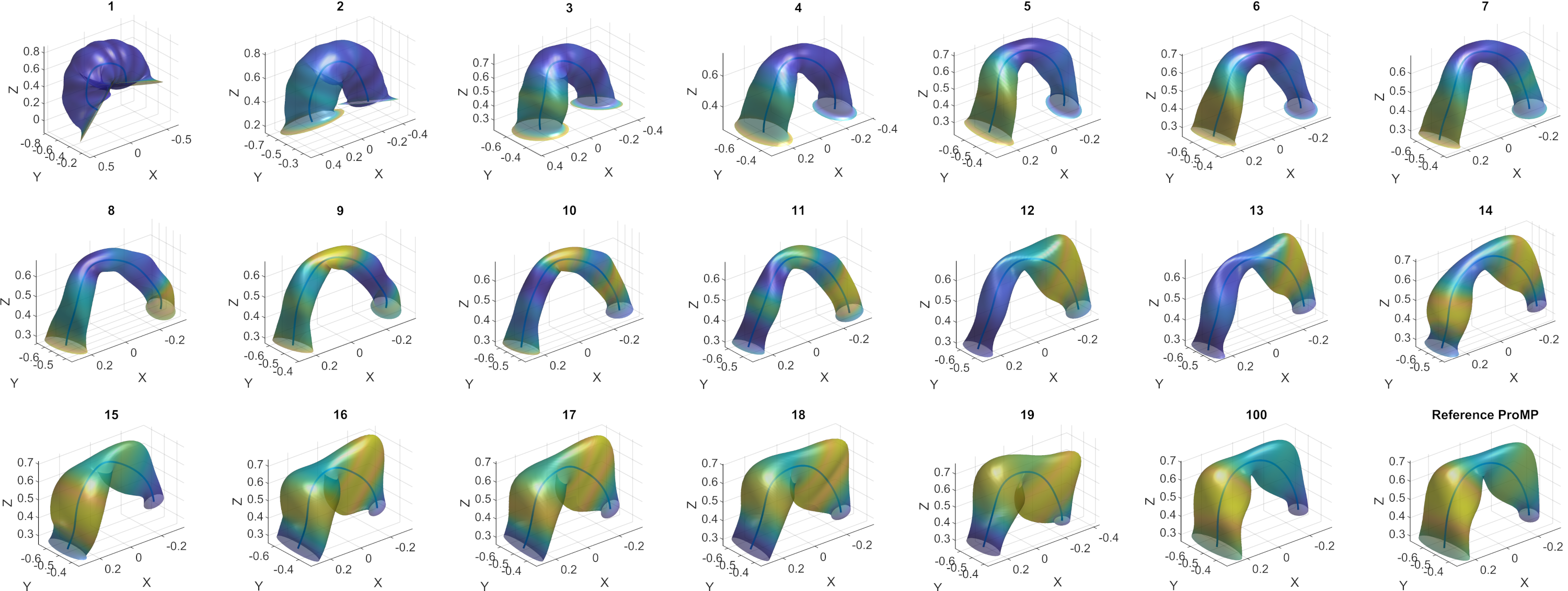}
	\caption{Evolution of the ProMP during incremental training on the first 19 demonstrations as well as the final ProMP after 100 demonstrations and the reference ProMP for comparison. The blue lines represent the mean $ \v{\mu}_w $ and the tubes the variance $ \v{\Sigma}_w $ in terms of two standard deviations. The demonstration number is shown on the top of each plot. During the first few demonstrations, the variance is strongly influenced by the initial value $ \v{\Sigma}_w = \I $. In course of training, the ProMP becomes increasingly similar to the reference.}
	\label{fig:incProgProMPs}
\end{figure*}

The previously described indicators are computed each time a new demonstration is added to the ProMP. Fig. \ref{fig:incProgMetrics} shows all four indicators plotted over the course of training. The Bhattacharyya distances reach an almost steady level after around 40 demonstrations. The relative errors computed from the Frobenius norms reach near steady levels immediately for the mean $ \v{\mu}_w $ and after about 15 demonstrations for the covariance $ \v{\Sigma}_w $. 		
To visualize the incremental training progress we show the evolution of the ProMP distribution during the training on the first 19 demonstration, as well as the final ProMP after 100 demonstrations and the reference ProMP in Fig.~\ref{fig:incProgProMPs}. 

\subsection{Adaptation to Changes in the Task} \label{ssec:ExpAdapt2Changes}
\begin{figure}[t]
	\centering
	\subfloat{\includegraphics[width=0.5\linewidth]{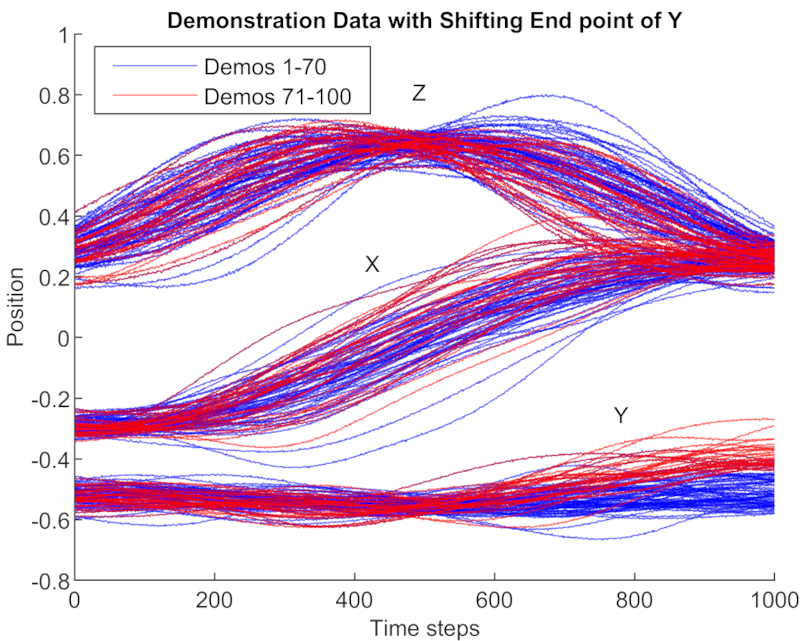}}
	\subfloat{\includegraphics[width=0.5\linewidth]{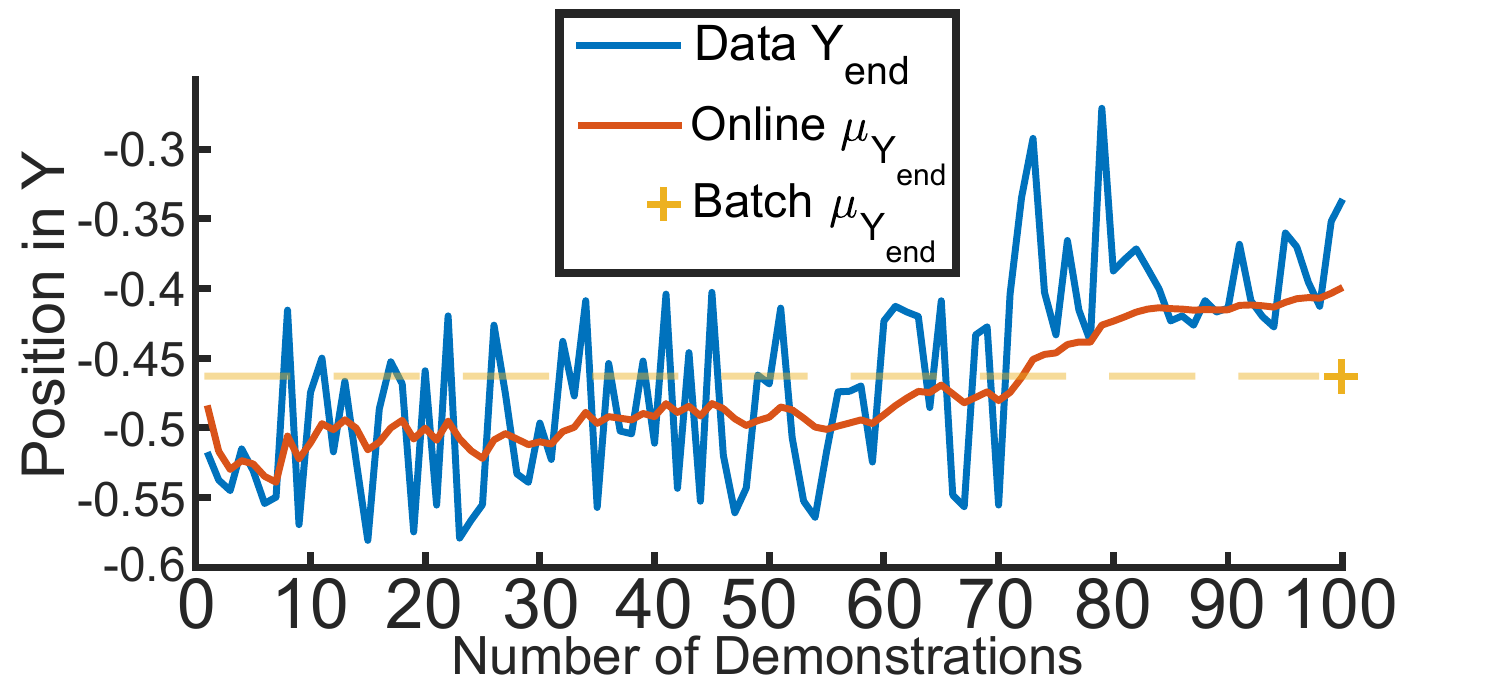}}
	\caption{\textbf{Left:} Data set with manipulated order of demonstrations to simulate a change in task constraints. In this case, the endpoint of the trajectory of the Y component is shifted to a new mean. The first $ 70 $ demonstrations are shown in \textbf{blue}, the last $ 30 $ demonstrations, representing the shift in the task constraints, are shown in \textbf{red}.
	\textbf{Right:} The endpoint of the trajectory of Y plotted over the number of demonstrations is shown in \textbf{Blue}. The shift after demonstration 70 is visible. The \textbf{Red} line shows the endpoint of the mean trajectory of the Y component trained with our incremental algorithm. The mean follows the shift in the data. The \textbf{Yellow} cross shows the endpoint of the mean trajectory of the Y component trained with the batch EM algorithm. Due to the batch nature of the algorithm, the mean is only computed once after all demonstrations are added. The batch-mean roughly represents the mean of the entire set, or history, of demonstrations and has a rather large distance to the data from demonstration 70 to 100.}
	\label{fig:exp2dataset}
\end{figure}

\begin{figure}[t]
	\centering
	\includegraphics[width=1\linewidth]{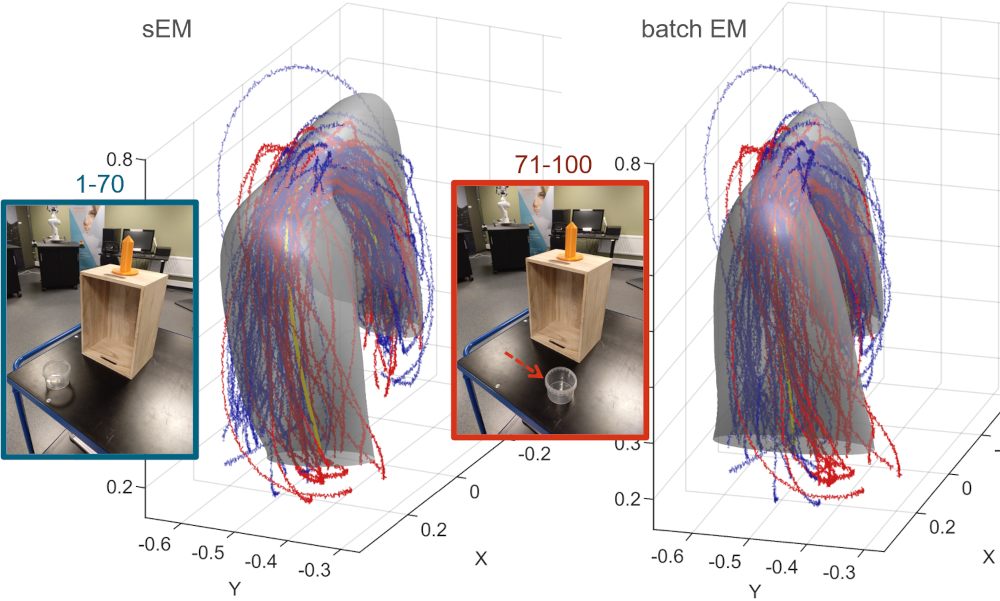}	
	\caption{Comparison of the ProMPs trained in presence of a shift in the task constraints. \textbf{Left:} Training result of our online algorithm. \textbf{Right:} Training result of the batch EM algorithm.
	The \textbf{blue} lines represent the first 70 demonstrations. The \textbf{red} lines represent the last 30 demonstrations with a shifted mean in Y direction at the end of the trajectory.
	The ProMP mean is shown as a \textbf{yellow} line, the variance (two standard deviations) as the \textbf{grey} tube.
	The incrementally trained ProMP \textbf{left} represents mostly the recent demonstrations \textbf{red}, while the ProMP trained with batch EM represents the entire distribution of demonstrations \textbf{blue} and \textbf{red}.}
	\label{fig:splitDataComp}
\end{figure}

In this experiment, we demonstrate how our online algorithm adapts the ProMP parameters to shifts in the task constraints/demonstration data. We use the same set of demonstrations as before, but manipulate its order as follows: We sort the demonstration data according to the endpoint (mean of the last five time steps) of the Y coordinate at cup A (Fig.~\ref{fig:lbdcomb4}). We then separate $ 30 $ demonstrations with the largest endpoint values from the remaining $ 70 $ demonstrations. The order within both subsets is randomized again. Finally, the subset containing the demonstrations with larger endpoint values is placed after the subset containing the remaining demonstrations. The resulting set of demonstrations has two distinct mean endpoints for Y; one mean for the first 70 demonstrations and one mean for the last 30 demonstrations. See Fig.~\ref{fig:exp2dataset} for clarification. The idea behind this arrangement is to simulate a long training phase with similar task constraints in which the training algorithms converge to a set of parameters. This is then followed by a change in the task constraints, represented by a shifted mean for the end of the movement in the Y direction. In the pick-and-place task in Fig~\ref{fig:lbdcomb4}, this shift would correspond to moving cup A to the right, resulting in a new "place"-position in the task. We claim that our online algorithm is well suited to capture such changes in a task constraints.
We compare our online algorithm with a step size reduction power $ \beta = 0.6 $ to the batch algorithm MAP with EM (5 iterations), which had the best performance in the comparison of the training algorithms in section~\ref{sec:ComparisonAlgorithms}.
Fig.~\ref{fig:splitDataComp} shows the resulting ProMPs of both algorithms together with the demonstration data.
The ProMP trained with sEM is adapted to the most recent data (red lines) by means of both mean and variance.
The ProMP trained with batch EM, however, represents the entire set of demonstrations, with a mean in the middle of the first (blue) and second (red) subset of demonstrations. The variance contains the entire set of demonstrations.
The mean endpoints of the Y component during the course of training are shown on the right in Fig.~\ref{fig:exp2dataset}. In case of sEM, the adaptation of the ProMP mean to the new endpoint is clearly visible.

\section{PROOF OF CONCEPT: ADAPTATION TO CHANGES IN THE TASK}\label{sec:ProofOfConcept}
\begin{figure}[t]
	\centering
	\includegraphics[width=1\linewidth]{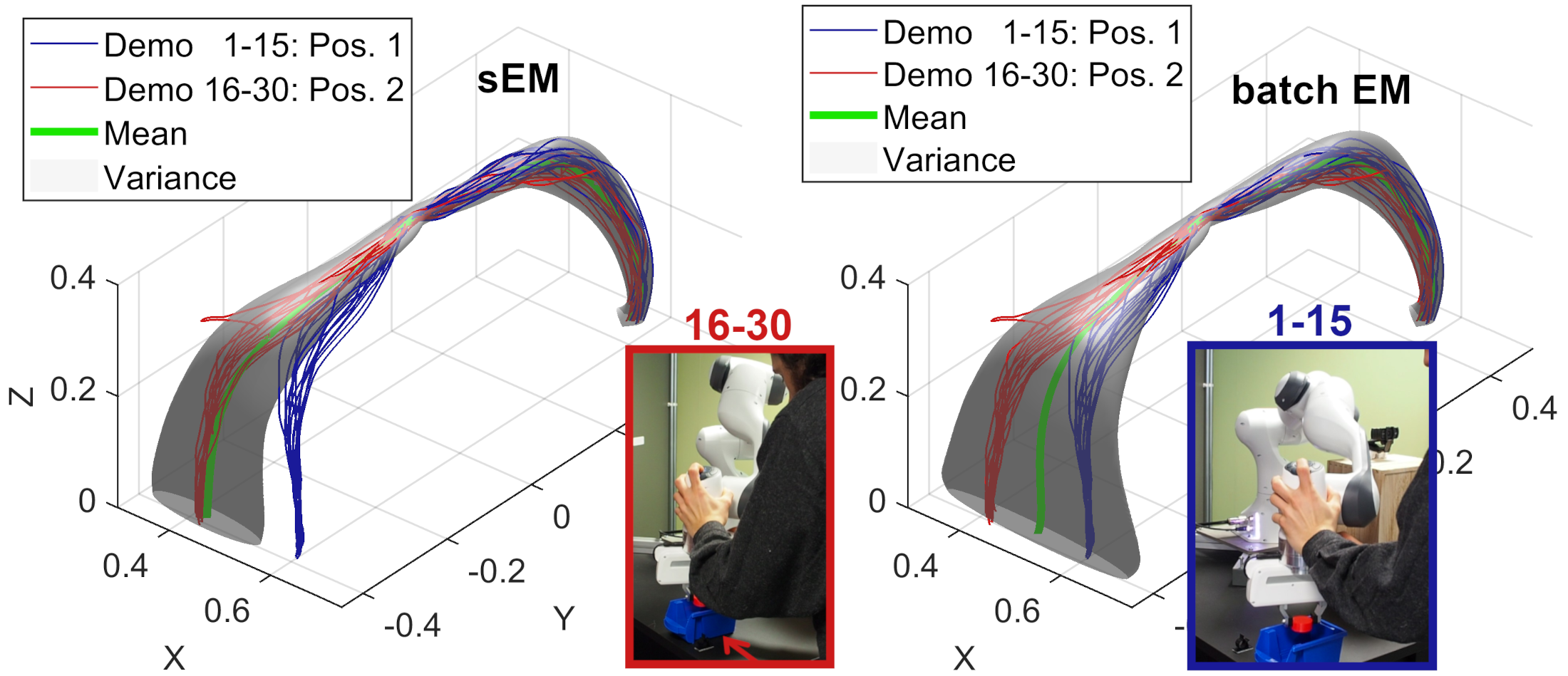}
	\caption{Proof of Concept on Franka Emika Panda robot. Comparison of ProMPs trained during a pick-and-place task with a change in the task constraints. \textbf{Left:} Training result of our online algorithm. \textbf{Right:} Training result of the batch EM algorithm. Axis units are meters.
	The \textbf{blue} lines represent the first 15 demonstrations. The \textbf{red} lines represent the last 15 demonstrations with a shifted mean in X direction at the end of the trajectory.
	The ProMP mean is shown as a \textbf{green} line, the variance (two standard deviations) as the \textbf{grey} tube.
	The mean \textbf{green} of the incrementally trained \textbf{left} is shifted to the most recent demonstrations \textbf{red}. The uncertainty about this new mean is still increased, indicated by the relatively large variance \textbf{grey}.
	The ProMP trained with batch EM represents the entire distribution of demonstrations \textbf{blue} and \textbf{red}, with a mean \textbf{green} that ends between the two "place"-positions.}
	\label{fig:pandaadaptcomp}
\end{figure}

To show the adaptation capabilities of our algorithm on unprocessed robot data, we conducted a similar pick-and-place experiment as in section~\ref{ssec:ExpAdapt2Changes} on a Franka Emika Panda manipulator.
The task was to pick up a part at a fixed position, move it precisely past an inspection camera and place it in a small container.
To change the task constraints slightly, the container was moved about 20cm to a new "place"-position after 15 demonstrations.
Then, another 15 demonstrations for the new "place"-position were provided.
We applied our incremental algorithm and the batch algorithm MAP with EM on the 30 demonstrations, as described in section~\ref{ssec:ExpAdapt2Changes}.
The resulting ProMPs are shown in Fig.~\ref{fig:pandaadaptcomp}.
The mean of the incrementally trained ProMP is shifted to the the new "place"-position shown on the left of Fig.~\ref{fig:pandaadaptcomp}.
In contrast, the ProMP trained with batch EM, shown on the right in Fig.~\ref{fig:pandaadaptcomp}, represents the entire distribution of demonstrations, with a mean that ends between the initial and the final "place"-position and a variance that contains both the "place"-positions.

\section{DISCUSSION}
In the following section, we discuss the results from the three experiments in section \ref{sec:Experiments} and relate them to the general context of learning motor skills incrementally in pHRC.

\subsection{Discussion on Training Algorithm Performance}
Looking at the results in Table~\ref{tab:DistancesToRef}, we see that all algorithms lead to comparable results. 
As expected, the algorithms doing MAP estimation perform better in terms of the numerical stability of the covariance matrix, indicated by a lower condition number. Similar results are reported in \cite{GomezGonzalez.2020}.
The improved numerical stability is a consequence of the regularization through the prior parameter distribution used to compute the maximum a-posteriori estimates. MAP estimation with EM performs best of all algorithms in terms of both Bhattacharyya distance and condition number of the covariance matrix.
Our incremental algorithm (MAP with sEM) has the largest Bhattacharyya distance to the reference. However, if we let it pass the data five times, as the batch EM algorithms, the incremental algorithm performs second best. The online learning progress comes at the expense of a slight loss in accuracy of the final parameter estimates. Given that the Bhattacharyya distance of our online algorithm is in the same range as the distances of the other algorithms, and if we compare the plot of the final ProMP distribution trained with the online algorithm to the reference ProMP in Fig.~\ref{fig:incProgProMPs}, this loss in accuracy seems to be acceptable. This last statement is especially true in the light of the cooperative learning settings we aim for - where an incremental training progress is essential.

\subsection{Discussion on Incremental Training Progress}
Considering the evolution of the ProMP distribution in Fig.~\ref{fig:incProgProMPs}, the variance structure of the first few ProMPs contains not much of the characteristics of the final ProMP and is not descriptive of the task. For example, the via-point in the middle of the movement that has to be passed with high accuracy is not visible.
After around 15 demonstrations, the ProMP shows good correspondence to the final ProMP trained with all 100 demonstrations.
This correspondence after 15 demonstrations is also reflected in the relative error of the Frobenius norm $ E_F $ of the covariance matrix $ \v{\Sigma}_w $ shown as a red line in the top right plot of Fig.~\ref{fig:incProgMetrics}.
Apart from the relative error, also the dynamics in the covariance structure indicated by the rotation of the first principal component appear to settle after about 15 demonstrations.
The matrix condition number (Fig.~\ref{fig:incProgMetrics}, bottom left) is actually the lowest, at some points even lower than the reference, during the beginning of the training. This observation can be explained by the initially strong influence of the initial value for $ \v{\Sigma}_w $ which is chosen to be the identity matrix $ \I $ with a matrix condition number $ \log\kappa(\I) = 0 $, and the block diagonal prior distribution for $ \v{\Sigma}_w $. The block diagonal prior suppresses the correlation terms between the the X,Y and Z components of the movement, since it implies the off-diagonal blocks of $ \v{\Sigma}_w $ to be zero.
It follows, that, until the influence of the prior is overruled by the data, there is less collinearity in $ \v{\Sigma}_w $ which could make the matrix near-singular and lead to higher condition numbers.

The variance at the beginning of training is roughly a symmetric tube around the mean which reflects the strong influence of the initial value for the covariance matrix $ \v{\Sigma}_w = \I $.
This symmetric tube is acceptable at early stages of training since it simply reflects equal uncertainty in all directions perpendicular to the mean, where the amount of initial uncertainty could be controlled by a scaling factor $ \v{\Sigma}_w = \sigma_{Init.} \I $.
However, we suggest that the early variance structure can in the future be improved by use of prior parameter distributions inspired by features of human motion. 

The estimates of the ProMP mean $ \v{\mu}_w $ appear to be stable throughout the course of training when we consider its low relative error computed from the Frobenius norm (blue line in Fig.~\ref{fig:incProgMetrics} top right). Reaching a near-steady value not before 40 demonstrations, the Bhattacharyya distance (Fig.~\ref{fig:incProgMetrics} top left) indicates slower convergence than the relative errors of the mean and the covariance. However, it is currently unclear how much the Bhattacharyya distance is correlated with the applicability of a ProMP in practice. This relation has to be determined in further experiments involving actual robots.

\subsection{Discussion on Adaptation to Changes in the Task}
The experiment in section \ref{ssec:ExpAdapt2Changes} had the purpose to demonstrate how our online algorithm can adapt a ProMP to changes in the task/demonstrations due to the built-in forgetting factor.
Looking at the right side of Fig. \ref{fig:exp2dataset}, it is obvious that the incremental learning algorithm adapts the endpoint of the mean trajectory (red line) such as to follow the distribution of the recent demonstrations. This includes the period from demonstration 70 to 100, where the simulated change in the task is happening.
Applying the batch EM algorithm, on the other hand, yields a mean endpoint (yellow cross) based on the entire set of demonstrations. This mean endpoint has a rather large distance to the data after the change in the task.
The same conclusions can be made by looking at the entire ProMP distributions shown in Fig.~\ref{fig:splitDataComp}.
On the left, the ProMP trained with our online algorithm has adjusted in terms of mean and variance to the recent data (red) representing the change in the task.
On the right, the ProMP trained with the batch algorithm is not particularly adapted to the change in the task, but represents the entire data from before and after the change.

The results from the proof of concept in section~\ref{sec:ProofOfConcept} confirm the results of this experiment on unprocessed robot data.
The incrementally trained ProMP is adapted to the shift in the task constraints (Fig.~\ref{fig:pandaadaptcomp}, left). The adaptation is especially visible for the ProMP mean.
After the change in the task, the variance at the end of the demonstrations is relatively large compared to the actual spread of the data.
We would expect the variance to further decrease if more demonstrations after the change in the task were provided and when more of the information of the first position is forgotten. 
This can be seen in Fig.~\ref{fig:splitDataComp}, where 30 demonstrations after the change were provided and the variance represents the spread of the data after the change well.
The ProMP trained with the batch algorithm (Fig.~\ref{fig:pandaadaptcomp}, left) has a mean that ends between the initial and final "place"-position, which would be a suboptimal reference trajectory to execute this pick-and-place movement on a robot.

We have demonstrated that the forgetting factor in our online algorithm has the desired effect of adapting ProMPs during training. Hence, the basic idea of our learning framework in that the human can shape the robot's motor skill by giving corrective demonstrations can now be realized with our online algorithm. The results imply that our algorithm can incorporate a learning curve of the human as well as changes in the task constraints - both of which have to be expected for HRC in an unstructured environment.

\section{CONCLUSIONS AND FUTURE WORK}
In this paper, we proposed a new algorithm for the incremental learning of ProMPs from human demonstration.
The experiments indicate that our algorithm is suited for incremental learning and shaping of motor skills in pHRC.
In our ongoing research, we investigate how we can aid the estimation of the ProMP parameters in early stages of training with prior parameter distributions based on features of human motor control.

In our future work, we plan to merge the learning algorithm proposed in this paper with an adaptive compliance controller to implement the incremental learning framework suggested in Fig.~\ref{fig:IncLearningScheme}.
The underlying idea is to control the interaction forces between human and robot dependent on the variance of the ProMP and so gradually increase the robot's contribution to the task.


\section*{CRediT AND ACKNOWLEDGMENT}
{Daniel Schäle}: \textit{Conceptualization, Methodology, Investigation, Software, Writing - original draft.}
{Martin F. Stoelen}: \textit{Conceptualization, Writing - review \& editing.}
{Erik Kyrkjebø}: \textit{Conceptualization, Funding acquisition, Writing - review \& editing.}
We thank {Johannes Møgster} for the valuable discussions and his help to realize the lab experiments.


\bibliographystyle{./IEEEtran}
\bibliography{./IEEEabrv,./lit}

\end{document}